\documentclass[10pt,twocolumn,letterpaper]{article}

\usepackage{cvpr}
\usepackage{times}
\usepackage{epsfig}
\usepackage{rotating, graphicx}
\usepackage{amsmath}
\usepackage{amssymb}
\usepackage{multirow}
\usepackage{float}

\usepackage{algorithmic}
\usepackage{algorithm}

\def\comment#1{{\color{red}#1}}
\def\cut#1{{\color{blue}#1}}
\def\comment#1{{\color{red}}}
\def\cut#1{{\color{blue}}}

\usepackage[pagebackref=true,breaklinks=true,letterpaper=true,colorlinks,bookmarks=false]{hyperref}

\cvprfinalcopy % *** Uncomment this line for the final submission

\def\cvprPaperID{722} % *** Enter the CVPR Paper ID here

\ifcvprfinal\fi
\begin{document}

%%%%%%%%% TITLE
\title{An Empirical Evaluation of Current Convolutional Architectures' Ability to Manage Nuisance Location and Scale Variability}

\author{Nikolaos Karianakis\\
{\tt\small nikarianakis@ucla.edu}\\
\and
Jingming Dong\\
{\tt\small dong@cs.ucla.edu}
\and
Stefano Soatto\\
{\tt\small soatto@ucla.edu}\\\\
\hspace{-9.5cm} UCLA Vision Lab, \; University of California, Los Angeles, CA 90095
}

\maketitle

%%%%%%%%% ABSTRACT
\begin{abstract}
We conduct an empirical study to test the ability of convolutional neural networks (CNNs) to reduce the effects of nuisance transformations 
of the input data,
such as location, scale and aspect ratio. We isolate factors by adopting a common convolutional architecture either deployed globally 
on the image 
to compute class posterior distributions, or restricted locally to compute class conditional distributions given location, scale and aspect ratios 
of bounding boxes determined by proposal heuristics. In theory, averaging the latter should yield inferior performance compared to proper 
marginalization. Yet empirical evidence suggests the converse, leading us to conclude that -- at the current level of complexity of convolutional 
architectures and scale of the data sets used to train them -- CNNs are not very effective at marginalizing nuisance variability. 
We also quantify 
the effects of context on the overall classification task and its impact on the performance of CNNs, and propose improved sampling techniques 
for heuristic proposal schemes that improve end-to-end performance to state-of-the-art levels. We test our hypothesis on a classification task 
using the ImageNet Challenge benchmark and on a wide-baseline matching task using the Oxford and Fischer's datasets.
\end{abstract}

%%%%%%%%% BODY TEXT

\section{Introduction}
\label{intro}

\comment{Motivation:}  Convolutional neural networks (CNNs) are the de-facto paragon for detecting the presence of objects in a scene, as portrayed by an image. CNNs are described as being ``approximately invariant'' to nuisance transformations such as planar translation, both by virtue of their architecture (the same operation is repeated at every location akin to a ``sliding window'' and is followed by local pooling) and by virtue of their approximation properties that, given sufficient parameters and transformed training data, could in principle yield discriminants that are insensitive to nuisance transformations of the data represented in the training set. In addition to planar translation, an object detector must manage variability due to scaling (possibly anisotropic along the coordinate axes, yielding different aspect ratios) and (partial) occlusion. Some nuisances are elements of a transformation group, \eg, the (anisotropic) location-scale group for the case of position, scale and aspect ratio of the object's support.\footnote{The region of the image the objects projects onto, often approximated by a bounding box.} The fact that convolutional architectures appear effective in classifying images as containing a given object regardless of its position, scale, and aspect ratio \cite{krizhevskySH12, simonyanZ14} suggests that the network can effectively manage such nuisance variability.

However, the quest for top performance in benchmark datasets has led researchers away from letting the CNN manage all nuisance variability. Instead, the image is first pre-processed to yield {\em proposals}, which are subsets of the image domain (bounding boxes) to be tested for the presence of a given class (Regions-with-CNN \cite{girshickDTM14}). Proposal mechanisms aim to remove nuisance variability due to position, scale and aspect ratio, leaving a ``Category CNN'' to classify the resulting bounding box as one of a number of classes it is trained with.
Put differently, rather than computing the {\em posterior} distribution\footnote{\label{foot2}One can think of the conditional distribution of a class $c$ given an image $x$, $p(c|x)$, as defined by a CNN, as the class posterior $\int_G p(c | x, g) dP(g | x)$ marginalized with respect to the nuisance group $G$. If the nuisances are known, one can use the class-conditionals $p(c | x, g_r)$ at each nuisance $g_r \in G$ in order to approximate $p(c|x)$ with a weighted average of conditionals, \ie, $p(c|x) \simeq \sum_r p(c|x,g_r) p(g_r | x)$.

When a CNN is tested on a proposal $r \subseteq x$ determined by a reference frame $x_r$, it computes $p(c | x_{|_r})$ ($x$ restricted to $r$), which is an approximation of $p(c | x, g_r)$. Then, explicit marginalization (assuming uniform weights) computes $\frac{1}{|r|}\sum_r p(c | x_{|_r})$ which is different from $\frac{1}{|r|}\sum_r p(c | x, g_r)$ which in turn is different from $\sum_r p(c | x, g_r)p(g_r | x)$. This approach is therefore, on average, a lower bound on proper marginalization, and the fact that it would outperform the direct computation of $p(c|x)$ is worth investigating empirically.} 
with nuisance transformations automatically marginalized, the CNN is used to compute the {\em conditional} distribution of classes given the data {\em and} a sample element that approximates the nuisance transformation, represented by a bounding box. If the goal is the nuisance itself (object support, as in {\em detection} \cite{deng09}) it can be found via maximum-likelihood ({\em max-out}) by selecting the bounding box that yields the highest probability of any class \cite{girshickDTM14, he14}. If the goal is the class regardless of the transformation (as in {\em categorization} \cite{deng09}), the nuisance can be approximately {\em marginalized out} by averaging the conditional distributions with respect to an estimation of the nuisance transformations${}^{\ref{foot2}}$.

Now, if a CNN was an effective way of computing the marginals with respect to nuisance variability, there would be no benefit in conditioning and averaging with respect to (inferred) nuisance samples. This is a direct corollary of the Data Processing Inequality (DPI, Theorem 2.8.1 in \cite{coverT12}). Proposals are subsets of the whole image, so in theory less informative even after accounting for resolution/sampling artifacts (Fig.~\ref{fig_rim_size}). {\em A fortiori}, performance should further decrease if the conditioning mechanism is not very representative of the nuisance distribution, as is the case for most proposal schemes that produce bounding boxes based on adaptively downsampling a coarse discretization of the location-scale group \cite{hosangBDS15}. Class posteriors conditioned on such bounding boxes discard the image outside it, further limiting the ability of the network to leverage on side information, or ``context''. Should the converse be true, \ie, should averaging conditional distributions restricted to proposal regions outperform a CNN operating on the entire image, that would bring into question the ability of a CNN to marginalize nuisances such as translation and scaling or else go against the DPI. In this paper we test this hypothesis, aiming to answer to the question: {\em How effective are current CNNs to reduce the effects of nuisance transformations of the input data, such as location and scaling?}

To the best of our knowledge, this has never been done in the literature, despite the keen interest in understanding the properties of CNNs 
\cite{goodfellow2009measuring, goodfellow2014adversarial, nguyenYC15, simonyanVZ14, szegedyZSBEGF14, yosinski2014nips, zeilerF14}
following their empirical success. We are cognizant of the dangers of drawing sure conclusions from empirical evaluations, especially when they involve a myriad of parameters and exploit training sets that can exhibit biases. To this end, in Sect.~\ref{sect-expm} we describe a testing protocol that uses recognized existing modules, and keep all factors constant while testing each hypothesis.

\subsection{Contributions}

We first show that a baseline (AlexNet \cite{krizhevskySH12}) with single-model top-5 error of $19.96\%$ on ImageNet 2014 Classification slightly {\em decreases} in performance (to $20.41\%$) when constrained to the ground-truth bounding boxes (Table \ref{table_gt}). This may seem surprising at first, as it would appear to violate Theorem 2.6.5 of \cite{coverT12} (on average, conditioning on the true value of the nuisance transformation must reduce uncertainty in the classifier). However, note that the restriction to bounding boxes does not just condition on the location-scale group, but also on {\em visibility}, as the image outside the bounding box is ignored. Thus, {\em the slight decrease in performance measures the loss from discarding context by ignoring the image beyond the bounding box.} When we pad the true bounding boxes with a 10-pixel rim, we show that, conditioned on such ``ground-truth-with-context'' indeed does decrease the error as expected, to  $17.65\%$. In Fig.~\ref{fig_rim_size} we show the classification performance as a function of the rim size all the way to the whole image for AlexNet and VGG16 \cite{simonyanZ14}. A $25\%$ rim yields the lowest top-5 errors on the ImageNet validation set for both models. This also indicates that the context effectively leveraged by current CNN architectures is limited to a relatively small neighborhood of the object of interest.

The second contribution concerns the {\em proper sampling} of the nuisance group. If we interpret the CNN restricted to a bounding box as a function that maps samples of the location-scale group to class-conditional distributions, where the proposal mechanism {\em down-samples} the group, then classical sampling theory \cite{shannon01} teaches that we should retain {\em not} the value of the function at the samples, but its {\em local average}, a process known as {\em anti-aliasing}. Also in Table \ref{table_gt}, we show that simple uniform averaging of 4 and 8 samples of the isotropic {\em scale} group (leaving location and aspect ratio constant) reduces the error to $15.96\%$ and $14.43\%$ respectively. This is again unintuitive, as one expects that averaging conditional densities would produce less discriminative classifiers, but in line with recent developments concerning ``domain-size pooling'' \cite{dongS15}.

\begin{table*}[t]
\begin{center}
\begin{tabular}{| c || c | c | c | c | }
\hline Method                     & \multicolumn{2}{c |}{{AlexNet}} & \multicolumn{2}{c |}{{VGG16}} \\ \hline \hline
Whole image                       & \multicolumn{2}{c |}{{19.96}}   & \multicolumn{2}{c |}{{13.24}} \\ \hline
Ground-Truth Bounding Box (GT)    & \multicolumn{2}{c |}{{20.41}}   & \multicolumn{2}{c |}{{12.44}} \\ \hline \hline
                                   & Isotropically & Anisotropically & Isotropically & Anisotropically \\ \hline \hline
GT padded with 10 px            & 17.66 & 17.65 & 10.91 & 10.30                     \\ \hline
Ave-GT, 4 domain sizes (padded with [0,30] px)  & 15.96 & 16.00 &  9.65 &  8.90 \\ \hline
Ave-GT, 8 domain sizes (padded with [0,70] px)  & 14.43 & 14.22 &  8.66 &  7.84 \\ \hline
\end{tabular}
\end{center}
\caption{\sl \small AlexNet's and VGG16's top-5 error on the ImageNet 2014 classification challenge when the ground-truth localization is provided, compared to applying the model on the entire image. We pad the ground truth with various rim sizes both isotropically and anisotropically. Then we show how averaging the class posteriors performs when applying the network on concentric domain sizes around the ground truth.}
\label{table_gt}
\end{table*}

To test the effect of such anti-aliasing on a CNN absent the knowledge of ground truth object location, we follow the methodology and evaluation protocol of \cite{fischerDB14} to develop a domain-size pooled CNN and test it in their benchmark classification of wide-baseline correspondence of regions selected by a generic low-level detector (MSER \cite{matasCUP04}). Our third contribution is to show that this procedure improves the baseline CNN by  $5$--$15\%$ mean AP on standard benchmark datasets (Table \ref{table_matching} and 
Fig.~\ref{fig_matching} in Sect.~\ref{subsection-matching}).

Our fourth contribution goes towards answering the question set forth in the preamble: We consider two popular baselines (AlexNet and VGG16) that perform at 
the state-of-the-art in the ImageNet Classification challenge and introduce novel sampling and pruning methods, as well as an 
adaptively weighted marginalization based on the inverse R\'enyi entropy. Now, if {\em averaging} the conditional class posteriors obtained with various
sampling schemes should improve overall performance, that would imply that the \emph{implicit} ``marginalization'' performed by the CNN is inferior to that obtained by sampling the group, 
and averaging the resulting class conditionals.${}^{\ref{foot2}}$ This is indeed our observation, \eg, for VGG16, as we achieve an overall performance of $8.02\%$, 
compared to $13.24\%$ when using the whole image (Table~\ref{table_imagenet}).  There are, however, caveats to this answer, 
which we discuss in Sect.~\ref{sect-discussion}.

Our fifth contribution is to actually provide a method that performs at the state of the art in the ImageNet Classification challenge 
when using a single model. 
In Table \ref{table_imagenet} we provide various results and time complexity. We achieve a top-5 classification error 
of $15.82\%$ and $8.02\%$ for AlexNet and VGG16, compared to $17.55\%$ and $8.85\%$ error when they are tested with $150$ 
regularly sampled crops \cite{simonyanZ14}, which corresponds to $9.9\%$ and $9.4\%$ relative error reduction, respectively. 
Data augmentation techniques such as scale jittering and an ensemble of several models \cite{he15, simonyanZ14, szegedyLJSAEVR14} 
could be deployed along with our method.

The source code implementing our method and the scripts necessary to reproduce the evaluation are available at 
\url{http://vision.ucla.edu/~nick/proj/cnn_nuisances/}.

%In the Supp. Mat. we report additional evaluation on {\em detection} task, where the benefit of computing conditionals is already obvious and implicit in the choice of most of the top performers in detection challenges to adopt a proposal or sampling mechanism. There, the element of the group transformation (bounding box) is no longer the nuisance, but the object of interest. 

\subsection{Related work}

The literature on CNNs and their role in Computer Vision is rapidly evolving.
Attempts to understand the inner workings of CNNs are being conducted 
\cite{chatfieldSVZ14, goodfellow2009measuring, goodfellow2014adversarial, lee2014deeply, nguyenYC15, simonyanVZ14,  szegedyZSBEGF14, yosinski2014nips, zeilerF14}, 
along with theoretical analysis \cite{anselmiRP15, brunaM13, cohenW14, soattoC16} 
aimed at characterizing their representational properties. Such intense interest was 
sparked by the surprising performance of CNNs \cite{chatfieldSVZ14, donahue2014long, girshickDTM14, he15, krizhevskySH12, renHGS15, sermanetEZMFL13, simonyanZ14, szegedyLJSAEVR14} in Computer Vision benchmarks \cite{deng09,everinghamGWWZ10}, 
where many couple a proposal scheme \cite{alexeDF12, carreiraS12, chengZLT14, erhanSTA14, hosangBDS15, humayunLR14, krahenbuhlK14, manenGG13, rahtuKB11, uijlingsSGS13, zitnickD14} with a CNN. 
As our work relates to a vast body of work, we refer the reader to references in the papers that 
describe the benchmarks we adopt, namely \cite{chatfieldSVZ14}, \cite{krizhevskySH12} and \cite{simonyanZ14}.

Bilen et. al. \cite{bilenPT14} also explore the idea of introducing proposals in classification. However, their approach leverages 
on a significantly larger number of candidates and focuses on sophisticated classifiers and post-normalization of class posteriors. 
Our investigation targets selecting a very small subset of the most discriminative candidates among generic object proposals, 
while building on popular CNN models.

\vspace{-2mm}

\section{Experiments}
\label{sect-expm}

\subsection{Large-scale Image Classification}
\label{subsection-classification}

\paragraph{What if we trivialize location and scaling?}
First, we test the hypothesis that eliminating the nuisances of location and scaling by providing a bounding box for the object of interest will improve 
the classification accuracy. This is not a given, for restricting the network to operate on a bounding box prevents it from leveraging on context outside it. We use 
the AlexNet and VGG16 pretrained models, which are provided with the MatConvNet open source library 
\cite{vedaldi14matconvnet}, and test their top-1 and top-5 classification errors on the ImageNet 2014 classification
challenge \cite{deng09}. The validation set consists of $50,000$ images, where at each of them one ``salient" class is annotated a priori 
by a human. However, other ImageNet classes appear in many of the images, which can confound any classifier.

We test the classifier in various settings (Table \ref{table_gt}); first, by feeding the entire image to it and letting the classifier 
manage the nuisances. Then we test the ground-truth annotated bounding box and concentric regions that include it. 
We try both isotropic and anisotropic expansion of the ground-truth region. We observe similar behavior, which is also consistent for both models.

\begin{figure}[t]
\begin{minipage}{\linewidth}
  \hspace{-1.5mm}\includegraphics[width=8.4cm]{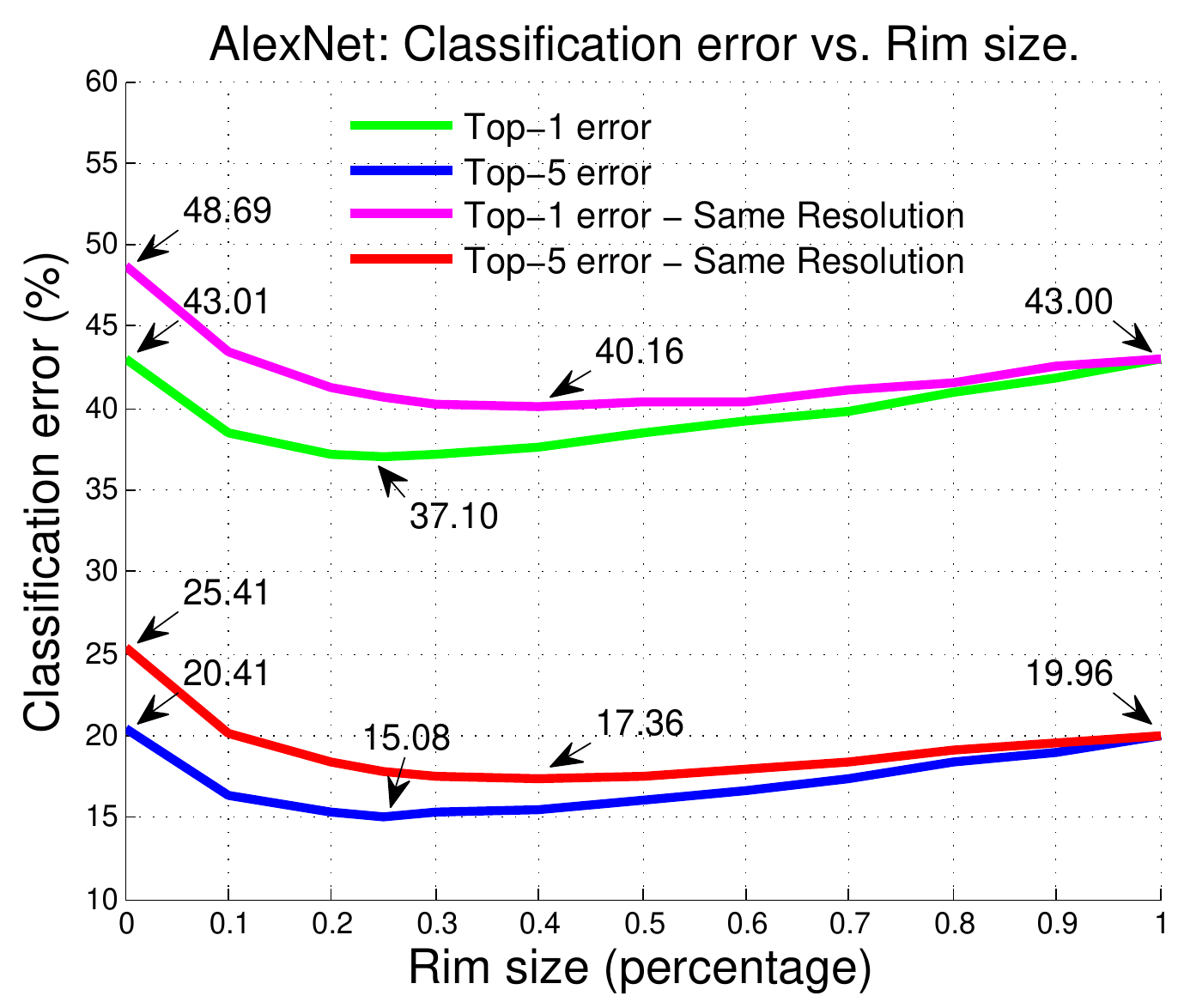}
\end{minipage}
\begin{minipage}{\linewidth}
  \hspace{-1.5mm}\includegraphics[width=8.4cm]{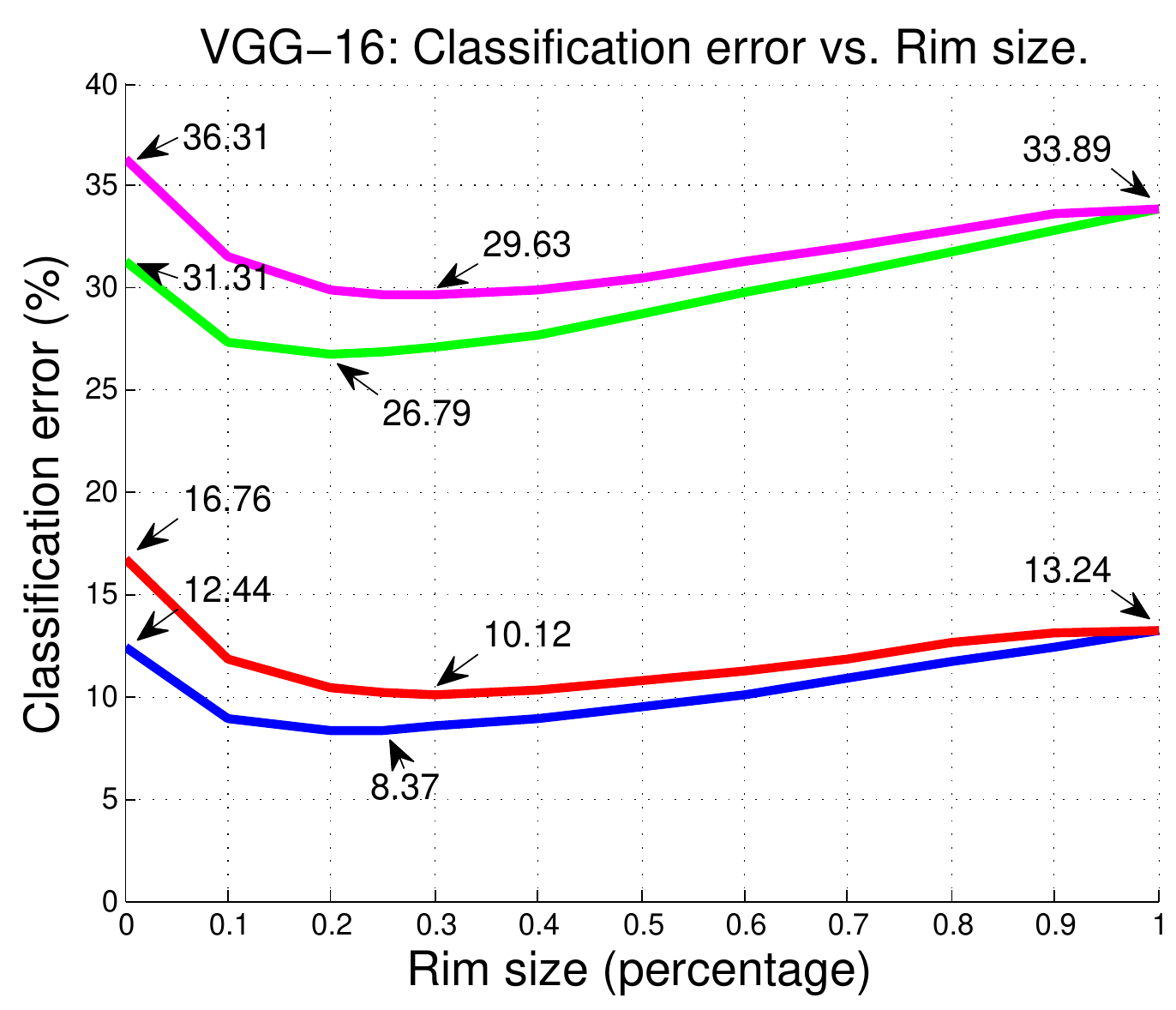}
\end{minipage}
\vspace{1mm}
\caption{The top-1 and top-5 classification errors in ImageNet 2014 as a function of the rim size for AlexNet (above) 
and VGG16 (below) architecture. A $0$ rim size corresponds to the ground-truth bounding box, while $1$ refers to the whole image.
A relatively small rim around the ground truth provides the best trade-off between informative context and clutter.}
\label{fig_rim_size}
\end{figure}

Only for AlexNet at Table \ref{table_gt} using the object's ground-truth support performs slightly worse than 
using the whole image. After we pad the object region with a $10$-pixel rim, 
the top-5 classification error decreases fast. However, there is a trade-off between context and clutter. 
Providing too much context has diminishing returns. In Fig.~\ref{fig_rim_size} we show how the errors vary 
as a function of the rim size around the object of interest. Performance starts dropping down when we add more than $25\%$ rim size.
This padding gives $15.08\%$ and $8.37\%$ top-5 error for AlexNet and VGG16, as opposed to $19.96\%$ and $13.24\%$ respectively, 
when classifying the whole image.

To ensure that this improvement is not due to downsampling, we repeat the experiment with fixed resolution for the whole image 
and every subregion. We achieve this by shrinking each region with the same downsampling factor that we apply to the whole image 
to pass to the CNN. Finally we rescale the downsampled region to the CNN input. These results appear with the label
``same resolution'' in Fig.~\ref{fig_rim_size}.
%They are characterized by a reduced but similar pattern of decrease.

Finally, we apply domain size average pooling on the class posterior (\ie, the network's softmax output layer) 
with $4$ and $8$ domain sizes that are concentric with the ground truth. 
The added rim has the declared size either at both dimensions (for the anisotropic case) or only along the minimum dimension 
(for the isotropic case), and it is uniformly sampled in the range $[0, 30]$ and $[0, 70]$, respectively. The latter one further reduces 
the top-5 error to $14.22\%$ for AlexNet, which is lower than any single domain size (\cf Fig.~\ref{fig_rim_size}). 
This suggests that explicitly marginalizing samples can be beneficial. Next we test whether the improvement stands 
when using object proposals.

\vspace{-3mm}

\paragraph{Introducing object proposals.}
We deploy a proposal algorithm to generate ``object" regions within the image. 
We use Edge Boxes \cite{zitnickD14}, which provide a good trade-off between recall and speed \cite{hosangBDS15}.

First, we decide the number of proposals which will provide a satisfactory cover for the majority of objects present in the dataset. 
In a single image we search for the highest Intersection over Union (IoU) overlap between the ground-truth region and any proposed sample 
and in turn we evaluate the network's performance on the most overlapping sample. 
We repeat this process for various number of proposals $N$ in a small subset of validation set and finally choose $N=80$, 
which provides a satisfactory trade-off between classification performance and computational cost.

\begin{figure}[t]
\begin{minipage}{0.50\linewidth}
\centerline{\includegraphics[width=4.1cm]{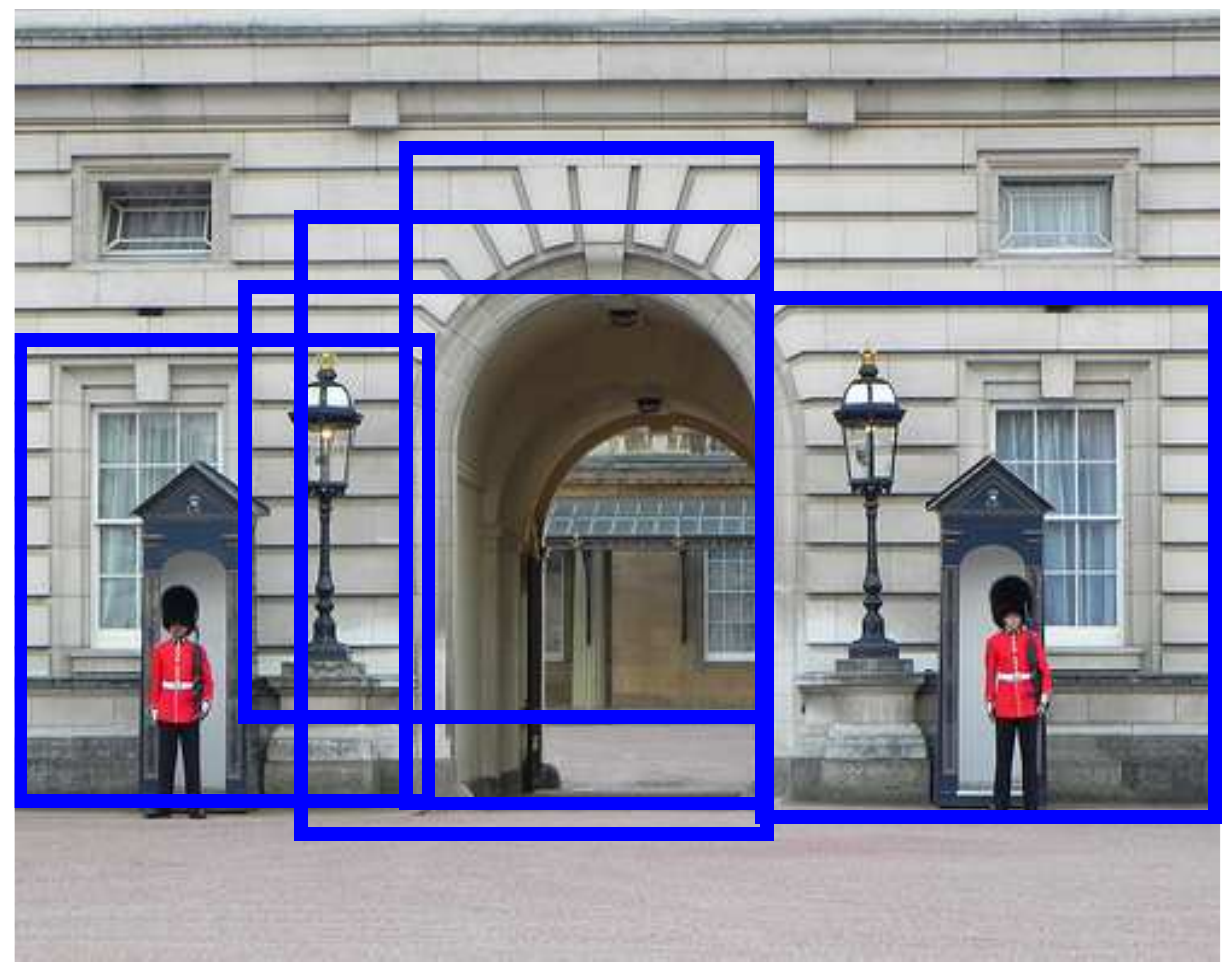}}
\end{minipage}%
\begin{minipage}{0.50\linewidth}
\centerline{\includegraphics[width=4.1cm]{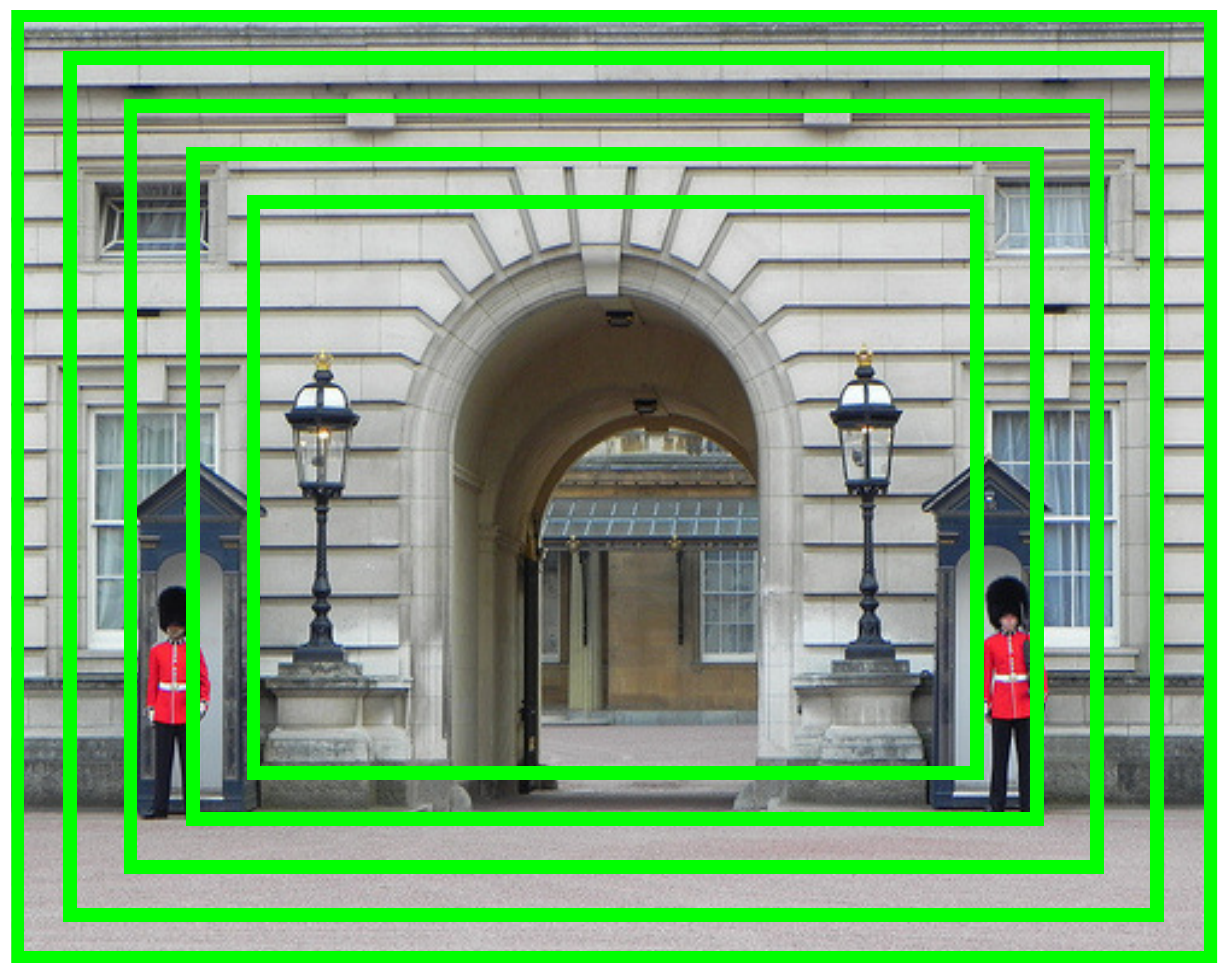}}
\end{minipage}%
\vspace{0.5mm}
\begin{minipage}{0.19\linewidth}
\centerline{\includegraphics[height=1.25cm]{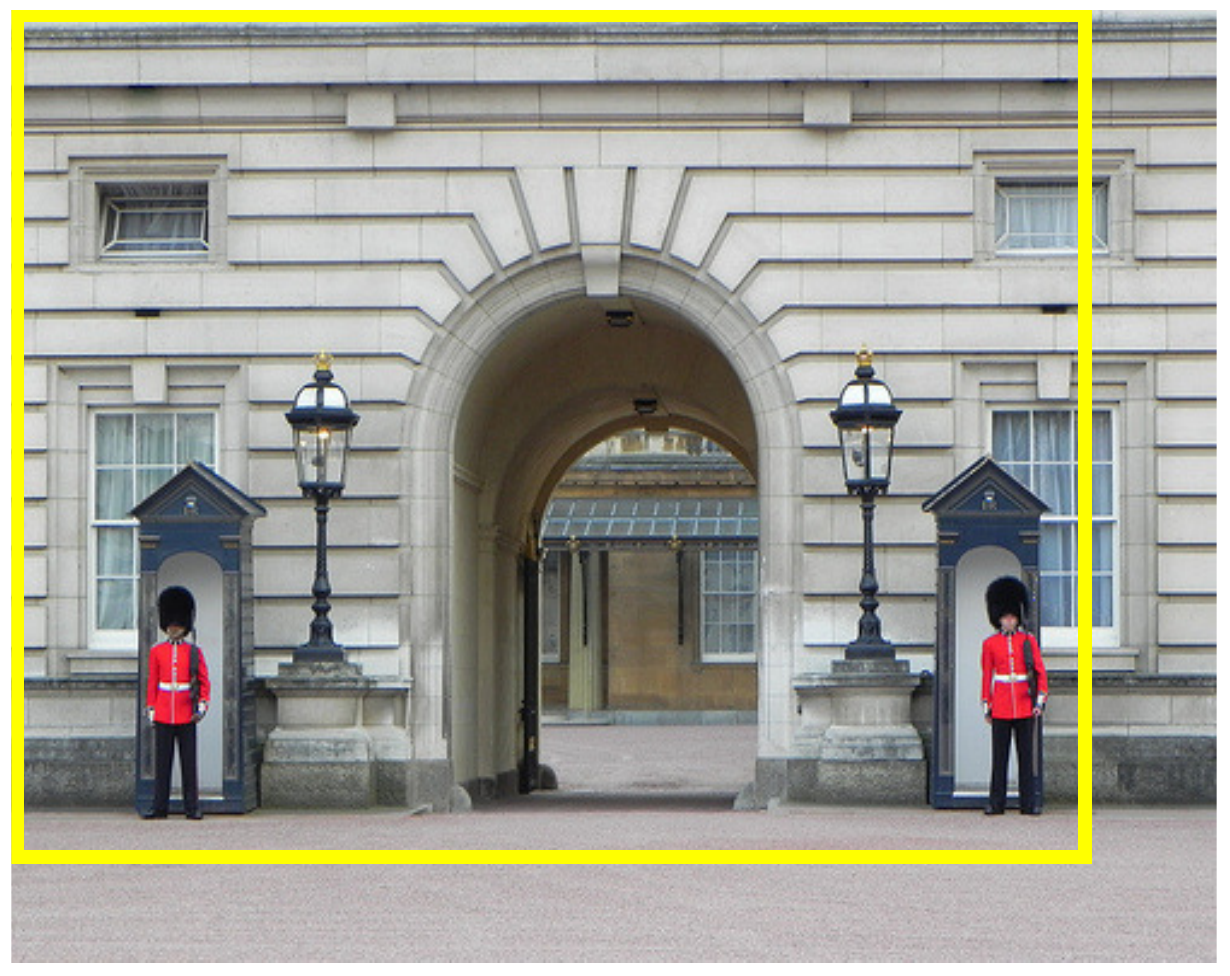}}
\end{minipage}
\hfill
\begin{minipage}{0.19\linewidth}
\centerline{\includegraphics[height=1.25cm]{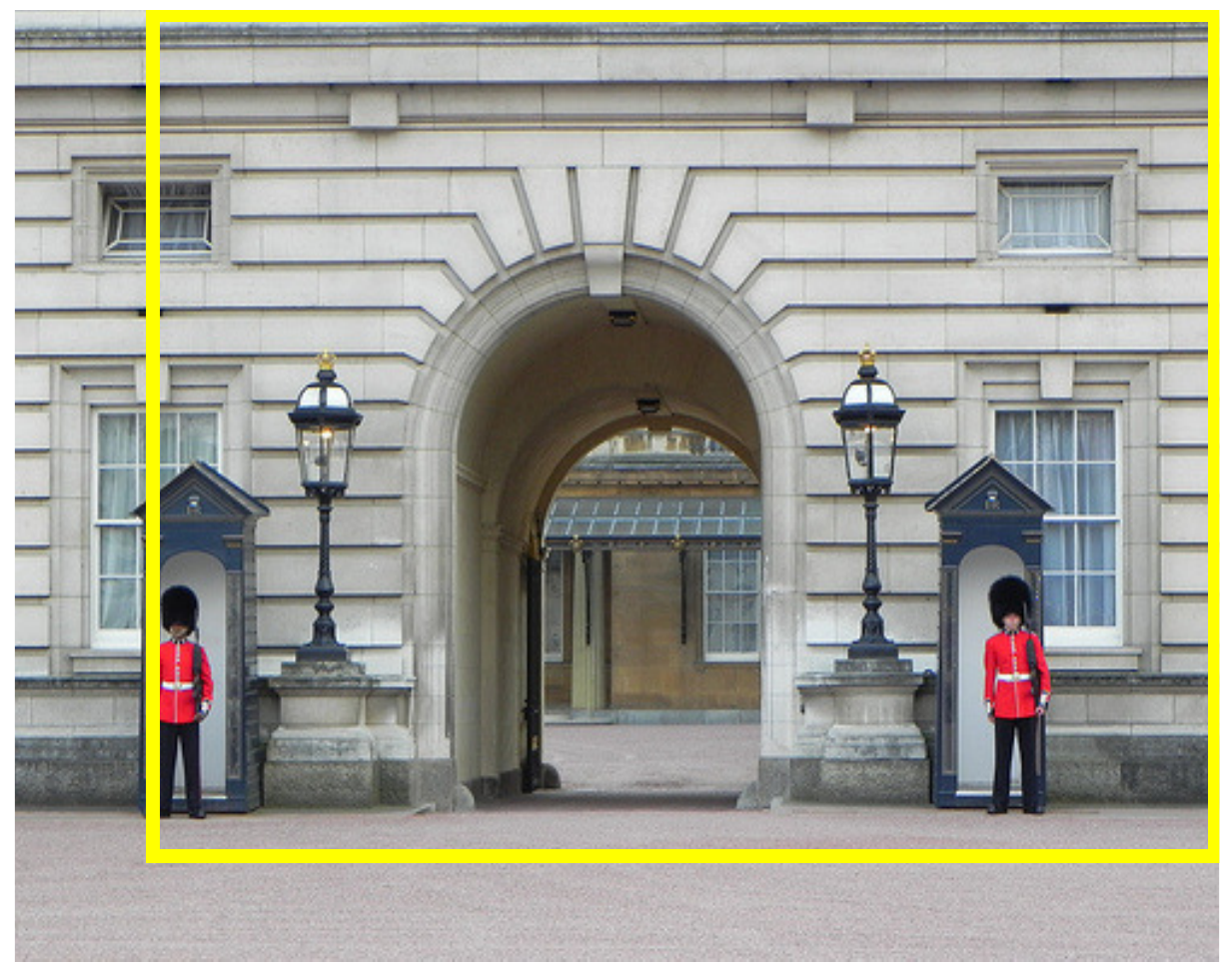}}
\end{minipage}
\hfill
\begin{minipage}{0.19\linewidth}
\centerline{\includegraphics[height=1.25cm]{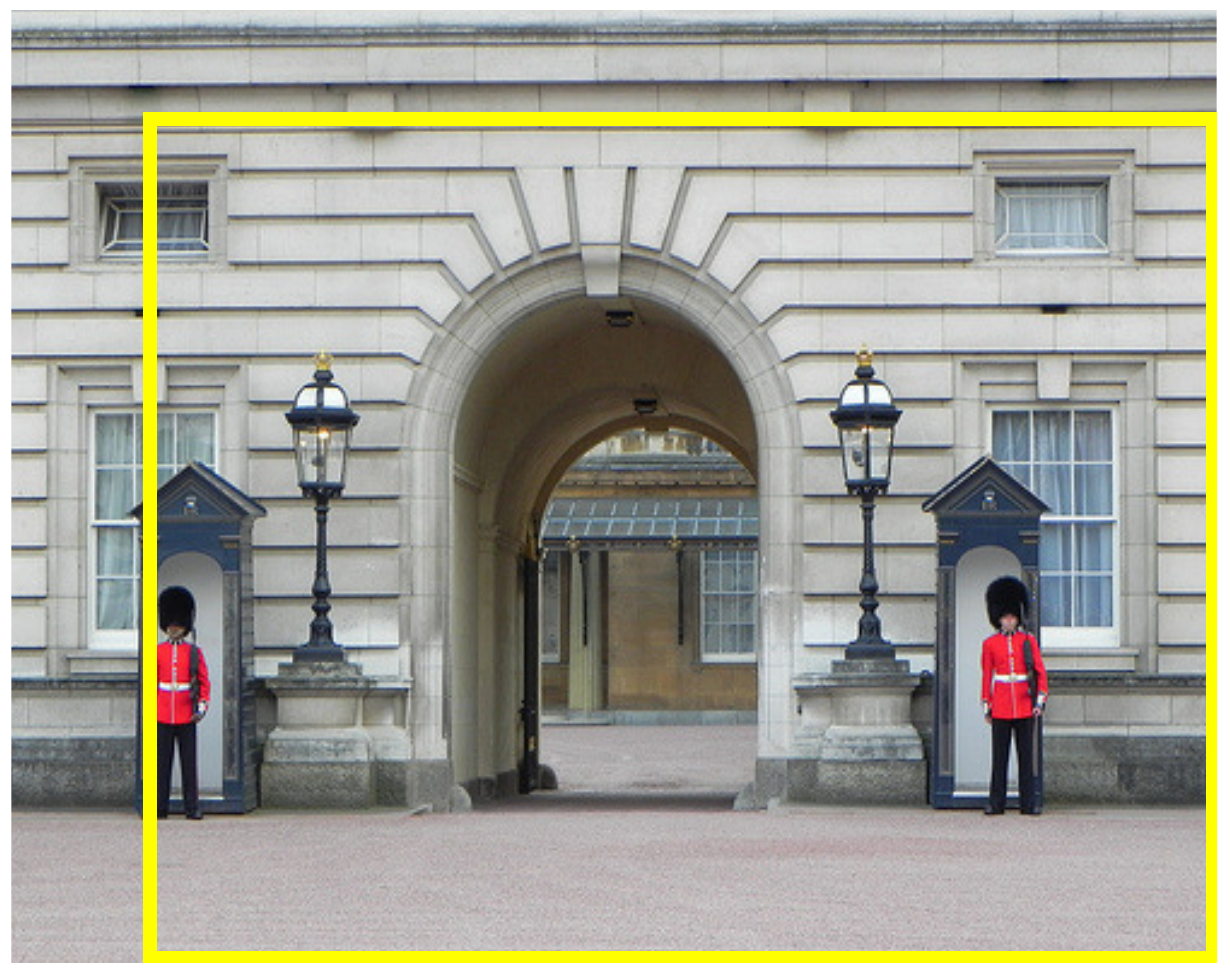}}
\end{minipage}
\hfill
\begin{minipage}{0.19\linewidth}
\centerline{\includegraphics[height=1.25cm]{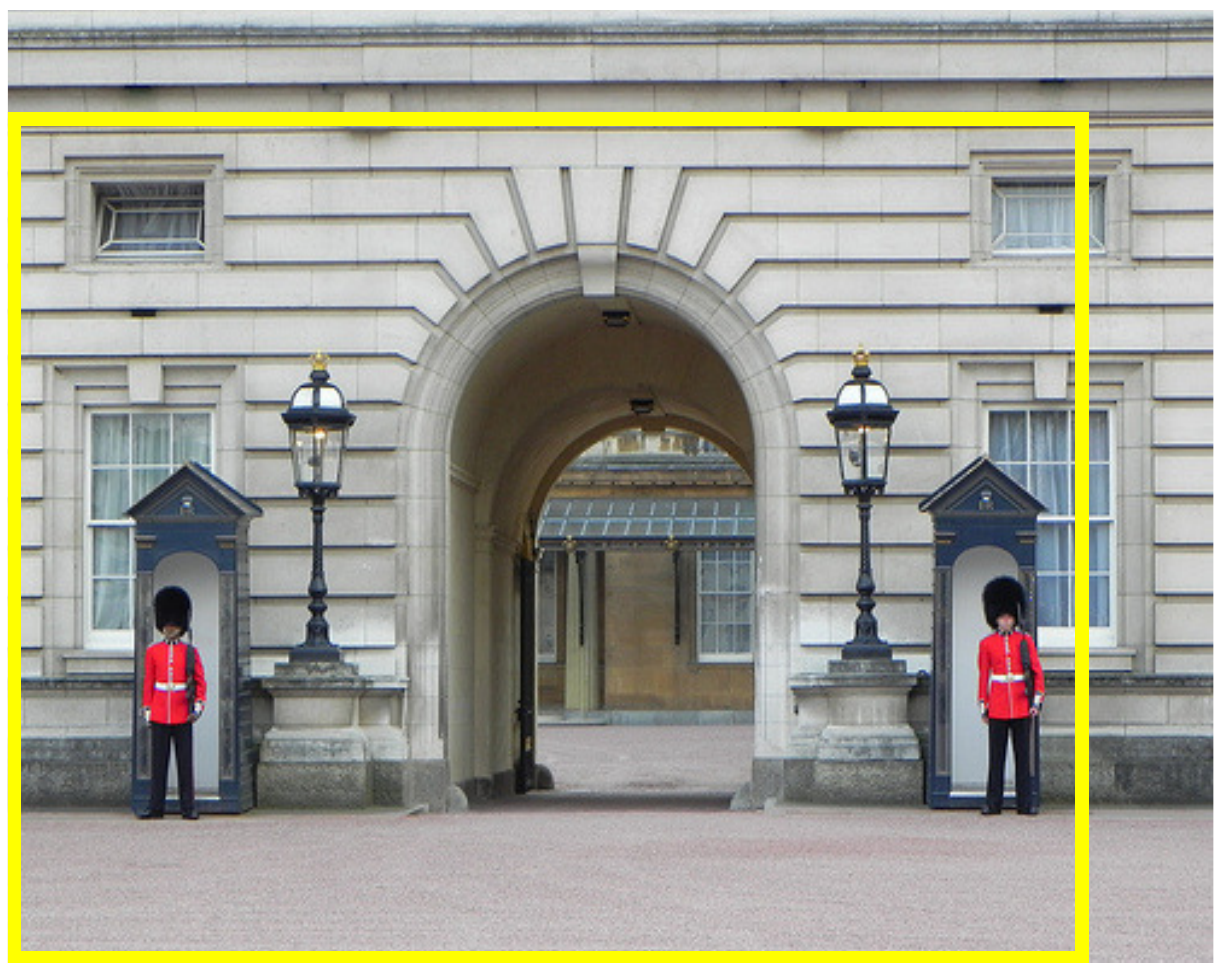}}
\end{minipage}
\hfill
\begin{minipage}{0.19\linewidth}
\centerline{\includegraphics[height=1.25cm]{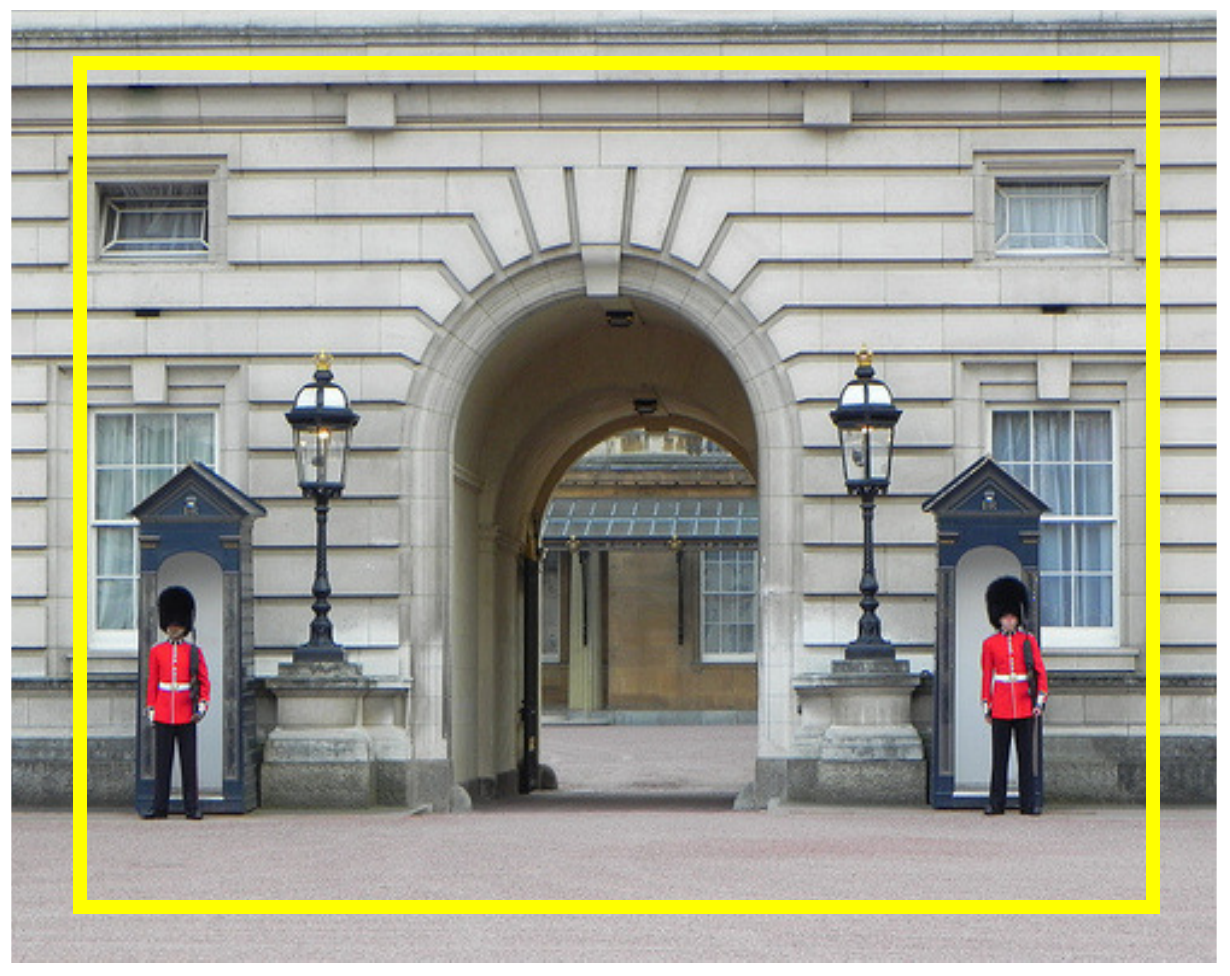}}
\end{minipage}
\vspace{2mm}
\caption{Visualizing different sampling strategies. 
Upper left:  Object proposals. Generic proposals using Edge Boxes \cite{zitnickD14}. 
Upper right: Concentric domain sizes are centered at the center of the image.
Below: Regular crops \cite{krizhevskySH12, simonyanZ14, szegedyLJSAEVR14}.}
\label{sampling_methods}
\end{figure}

Among the extracted proposals, we choose the most informative subset for our task, based on pruning criteria that we introduce later.
Next we discuss what other samples we use, which are also drawn in Fig.~\ref{sampling_methods}.

\vspace{-3mm}

\paragraph{Domain-size pooling and regular crops.}
We investigate the influence of domain-size pooling at test time both as stand-alone technique and as additional proposals 
for the final method which is described in Algorithm \ref{alg1}. We deploy domain-size aggregation of the network's class posterior over 
$D$ sizes that are uniformly sampled in the range $[r, 1]$, where $1$ is the normalized size of the original image. 
After parameter search, we choose $D=5$ and $r=0.6$. We use both the original and the horizontally flipped area, 
which gives $10$ samples in total.

Finally, we use standard data augmentation techniques from the literature. As customary, the image is isotropically rescaled to a 
predefined size, and then a predetermined selection of crops is extracted \cite{krizhevskySH12, simonyanZ14, szegedyLJSAEVR14}.

%or the network is applied densely and a class score map over the whole image is extracted \cite{sermanetEZMFL13, simonyanZ14}. We compare our method with the multi-crop strategies which have been shown to perform marginally better compared to dense processing \cite{simonyanZ14}.

%Krizhevsky et al. \cite{krizhevskySH12} rescale a test image to $256 \times 256$ and make predictions by extracting five $224 \times 224$ crops that are located at the four corners and the center, as well as their horizontal flips, and averaging the network's softmax output on these $10$ patches. They show that this strategy reduces the classification top-5 error from $18.3\%$ to $17.0\%$ on the ILSVRC $2010$ test set. Using a larger set of crops \cite{simonyanZ14, szegedyLJSAEVR14} can further improve the classification accuracy, as it results in a finer sampling of the input image. However, there is diminishing return as sampling more test crops can reduce the discriminability.

\vspace{-3mm}

\paragraph{Pruning samples.} 

Continuing to sample patches within the image has diminishing return in terms of discriminability, while including more background patches with noisy class posterior distribution. We adopt an information-theoretic criterion to filter the samples that we use for the subsequent approximate marginalization.

For each proposal $n \in N$ we evaluate the network and take the normalized softmax output $v^n \in {\mathbb R}^{\mathcal C}$, 
where $v_i^n \in [0,1], i=\{1,\ldots,{\mathcal C}\}$ and ${\mathcal C}=1,000$ on ILSVRC classification. 
The output is a set of non-negative numbers which sum up to 1. We can interpret the vector $v^n$ as a 
probability distribution on the discrete space of classes $\{1,\ldots,{\mathcal C}\}$ and compute the R\'enyi entropy as 
${\mathbb H_{\alpha}}(v^n) = \frac{1}{1-\alpha} log (\sum_{i=1}^{\mathcal C} (v_i^n)^{\alpha} )$.

\begin{figure}[t]
  \hspace{2mm}\includegraphics[width=8cm]{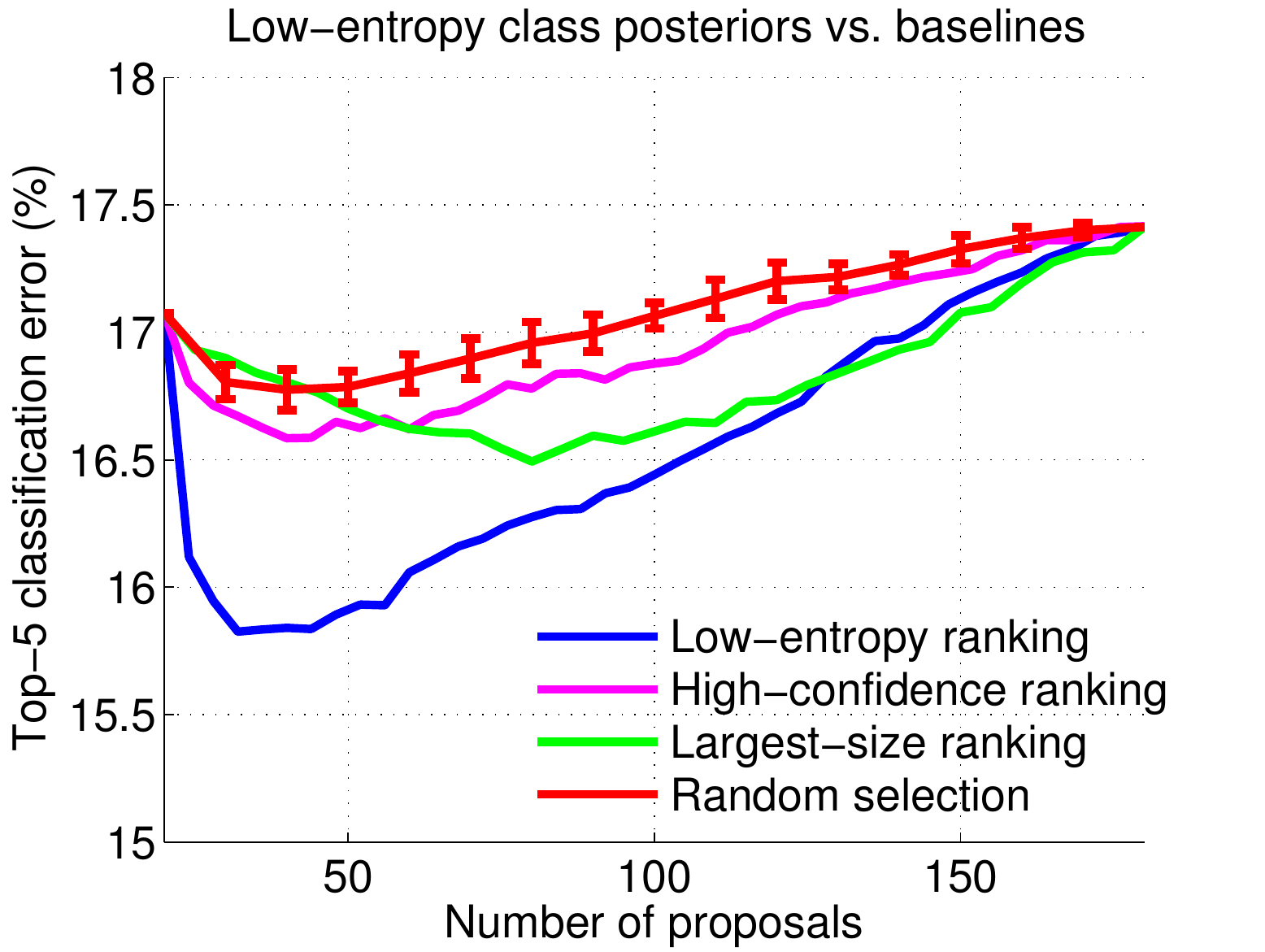}
  \caption{We show the top-5 error as a function of the number of proposals we average to produce the final posterior. 
Samples are generated with Algorithm \ref{alg1} and classified with AlexNet. 
The blue curve corresponds to selecting samples with the lowest-entropy posteriors. 
We compare our method with simple strategies such as random selection, ranking by largest-size or highest confidence of proposals.
The random sample selection was run $10$ times and we visualize the estimated $99.7\%$ confidence intervals as error-bars. 
Empirically, the discriminative power of the classifier increases when the samples are selected with the least entropy criterion.}
\label{entropy_plot}
\end{figure}

Our conjecture is that more discriminative class distributions tend to be more peaky with less ambiguity among the classes, and 
therefore lower entropy. In Fig.~\ref{entropy_plot} we show how selecting a subset of image patches whose class posterior has 
lower entropy improves classification performance.

We extract $N$ candidate object proposals\footnote{We introduce a prior encouraging the 
largest proposals among the ones that the standard setting in \cite{zitnickD14} would give. To this end, instead of directly extracting, for 
example, $N=80$ proposals, we generate $200$ and keep the $80$ largest ones (Algorithm \ref{alg1}).} \cite{zitnickD14} and 
evaluate the network for both the original candidates and their horizontal flips. Then we keep a small subset $E$, whose posterior distribution 
has the lowest entropy. We use R\'enyi entropy with relatively small powers ($\alpha=0.35$), as we found that it encourages 
selecting regions with more than one highly-confident candidate object. While the parameter $\alpha$ increases, the entropy is increasingly 
determined by the events of highest probability. Larger $\alpha$ would be more effective for images with a single object, which 
is not the case in most images in ILSVRC.

Finally we introduce a weighted average of the selected posteriors as $\sum_r p(c | x_{|_r}) p(x_{|_r})$, where $x_{|_r}$ is the support of sample 
$r$ and $p(x_{|_r})$ is the weight of its posterior${}^{\ref{foot2}}$. We try both uniform weights and weights proportional to the
inverse entropy of the posterior $p(c | x_{|_r})$. The latter is expected to perform better, as it naturally gives higher weight to 
the most discriminative samples.

%In the Supp. Material we provide more implementation details and we include comparisons with an additional pruning criterion and max-out inference.

%We introduce a prior encouraging the candidate proposals to contain a larger area than what the standard setting
%in \cite{zitnickD14} would give. To this end, instead of directly extracting $N=80$ proposals, we generate $200$ and keep the $80$ largest ones. 

%All regular and adaptive sampling components are summarized in Algorithm \ref{alg1} and are drawn in Fig. \ref{sampling_methods}. These $E$ proposals are classified by a Convolutional Neural Network and the multiple outputs are averaged element-wise in order to extract a single vector (of size $1,000 \times 1$ for Imagenet classification), which is our class posterior for the whole image.

\begin{algorithm}[b]
\caption{\sl \small Regular $\&$ adaptive sampling in classification.}
\label{alg1}
\begin{algorithmic}
\STATE
\STATE $\bullet$ \emph{Object proposals.} We extract several object proposals from the image $x$ (\eg, $200$ \emph{Edge Boxes} \cite{zitnickD14} 
and keep the $N$ largest ones). Among them we choose $E$ proposals whose class posterior has 
the lowest \emph{R\'enyi entropy} with parameter $\alpha$. After hyper-parameter search, we choose $N=80$, $E=12$ and $\alpha=0.35$.
\STATE $\bullet$ \emph{$D$ concentric} domain sizes around the center of $x$ (including their horizontal flip). We use $5$ sizes that are uniformly 
extracted in the normalized range $[ 0.6, 1 ]$, where $1$ corresponds to the whole image ($D=10$).
\STATE $\bullet$ \emph{$C$ crops.} Regular crops; \eg, $C=10$ or $C=50$ in $1$ or $3$ scales, as in \cite{krizhevskySH12, simonyanZ14, szegedyLJSAEVR14}.
\STATE $\bullet$ The class conditionals are approximated as $\sum_r p(c | x_{|_r}) p(x_{|_r})$, where $p(x_{|_r})$ is either uniform
or equals to the inverse entropy of the posterior $p(c | x_{|_r})$.
\end{algorithmic}
\end{algorithm}

\begin{table*}
\begin{center}
\begin{tabular}{| c | c | c || c | c | c | c | c | c | c | c |}
\hline \multicolumn{3}{|c||}{Method} & \multicolumn{3}{c|}{AlexNet} & \multicolumn{3}{c|}{VGG16} & \multirow{2}{*}{$\#eval$} & \multirow{2}{*}{$\#ave$} \\ \cline{1-9}
$\#$ crops & $\#$ sizes & $\#$ proposals  & top-1 & top-5 & t (s/im) & top-1 & top-5 & t (s/im) &    &    \\ \hline \hline
$-$             &  $D=1$     &     $-$          & 43.00 & 19.96 &  0.01     & 33.89 & 13.24 & 0.06    &  1 &  1 \\ \hline
$C=10$     &      $-$       &     $-$          & 41.50 & 18.69 & 0.06    & 27.55 &  9.29 & 0.48   & 10 & 10 \\ \hline
$C=50$     &      $-$       &     $-$           & 41.01 & 18.05 &  0.66    & 27.44 &  9.12 &   1.34      & 50 & 50 \\ \hline
$C=10\times3$ &  $-$    &    $-$           & 40.58 & 17.97 & 0.16 & 27.23 &  8.88 & 1.26 & 30 & 30 \\ \hline
$C=50\times3$ &  $-$    &     $-$          & {\bf 40.41} & {\bf 17.55} &  {\bf 0.82}  & {\bf 27.14} & {\bf 8.85} & {\bf 3.48}  & 150 & 150 \\ \hline \hline
$-$              &  $D=10$  &    $-$             & 40.00 & 17.86 & 0.08 & 28.16 &  9.46 & 0.60 & 10 & 10 \\ \hline
$C=10$      &  $D=10$  &   $-$              & 39.38 & 17.08 & 0.22  & 26.94 &  8.83 & 1.08 & 20 & 20 \\ \hline
$C=10\times3$ & $D=10$  &    $-$        & 39.36 & 17.07 & 0.46 & 26.76 &  8.68 & 1.88 & 40 & 40 \\ \hline
$-$              &  $-$           &    $E=40$    & 40.18 & 17.53 & \multirow{2}{*}{1.26} & 25.60 &  8.24 &  \multirow{2}{*}{3.02} & 160 & 40 \\ \cline{1-3}\cline{4-5}\cline{7-8}\cline{10-11}
$C=10$      &   $-$          &    $E=20$    & 38.91 & 16.63  & & 25.28 &  7.91 &   & 170 & 30 \\ \hline
$-$               & $D=10$    &   $E=12$    & 38.05 & 16.19 &  \multirow{2}{*}{1.34}  & 25.19 &  8.11 &  \multirow{2}{*}{4.38}  & 170 & 22 \\ \cline{1-3}\cline{4-5}\cline{7-8}\cline{10-11}
$C=10$       & $D=10$     &   $E=12$        & 37.69 & 15.83 &  & 25.11 & 8.01 &  & 180 & 32 \\ \hline
$C=10$       & $D=10$   &  $E=12$ (fast)  & 37.71 & 15.88 & 0.94 & 25.12 & 8.08 & 3.70 & 180 & 32 \\ \hline
$C=10$       & $D=10$   &   $E=12$ (W, fast)   & {\bf 37.57} & {\bf 15.82}  & {\bf 1.28} & {\bf 25.11} & {\bf 8.02} & {\bf 3.80} & 180 & 32 \\ \hline \hline
$C=10$       & $D=10$     &   $E=12$ ({\it test set})   & 37.417 & 16.018 & $-$     & 25.117 & 7.909 & $-$      & 180 & 32 \\ \hline
\end{tabular}
\end{center}
\caption{\small Top-1 and top-5 errors on the ImageNet 2014 classification challenge. The rows 2--5 include 
the common data augmentation strategies in the literature \cite{krizhevskySH12, simonyanZ14, szegedyLJSAEVR14} (\ie, regular sampling). 
The next three rows use concentric domain sizes that are uniformly sampled in the range $[0.6, 1]$ with $1$ being the normalized size of the original image 
(\cf Fig.~\ref{sampling_methods}). 
Finally, in the last seven rows, we introduce adaptive sampling, which consists of a data-driven object proposal algorithm \cite{zitnickD14} and an 
entropy criterion to select the most discriminative samples on the fly based on the extracted class posterior distribution. 
The last row shows results on the test set.
$\#eval$ stands for the number of samples that are evaluated for each method, while $\#ave$ is the number of samples 
that are eventually element-wise averaged to produce one single vector with class confidences. 
The previous top-reported with regular sampling and our results are shown in bold.}
\label{table_imagenet}
\end{table*}

\vspace{-3mm}

\paragraph{Comparisons.}

To compare various sampling and inference strategies, we use the AlexNet and VGG16 models. 
All classification results in Table \ref{table_imagenet} refer to the validation set of the ILSVRC 2014 \cite{deng09}, except for the last row which demonstrates results on the test set. On the rows $2$--$5$ we show the performance of popular multi-crop methods \cite{krizhevskySH12, simonyanZ14, szegedyLJSAEVR14}. Then we compare them with strategies that involve concentric domain sizes (rows $6$--$8$) and object proposals (rows $9$--$14$).

Before extracting the crops and in order to preserve the aspect ratio of each single image, we rescale it so that its minimum 
dimension is $256$. The proposals are extracted at the original image resolution and then they are 
rescaled anisotropically to fit the model's receptive field. Additionally, some multi-crop algorithms resize the image in $S$ different scales 
and then sample $C$ patches of fixed size $224 \times 224$ densely over the image. Szegedy et al. \cite{szegedyLJSAEVR14} use $S=4$ scales and $C=36$ 
crops per scale, which yields $144$ patches in all. Following the methodology from Simonyan et al. \cite{simonyanZ14}, 
it is comparable to deploy $S=3$ scales and extract $C=50$ crops per scale ($5\times5$ regular grid with flips), for a total of $150$ 
crops over $3$ scales (row $5$ in Table \ref{table_imagenet}).

The results, presented in Table \ref{table_imagenet}, indicate as expected that scale jittering at test time improves the 
classification performance for both 10-crop and 50-crop strategies. Additionally, the 50-crop strategy is better than the 10-crop strategy 
for both models. The results on row 5 in bold are the lowest errors that can be achieved with these specific single models\footnote{Specifically, 
we use the VGG16 model which is trained without scale 
jittering at training and appears on the first row of D area in Table 3 in \cite{simonyanZ14}.
Pre-trained models for both AlexNet and VGG16 are publicly available with the MatConvNet toolbox \cite{vedaldi14matconvnet}.
Simonyan et al.~in their evaluation with $50$ crops and $3$ scales report $8.6\%$ top-5 error on ImageNet 2014 validation. 
In contrast our implementation produces $8.85\%$, which can be attributed to using a different pre-trained model, as the initial weights are 
sampled from a zero-mean Gaussian distribution with standard deviation $0.01$ and there might also be minor differences in the training process.} using only regular crops.

Then we present our methods and observe that using the AlexNet network with $D=10$ concentric domain sizes 
outperforms most multi-crop algorithms even if it only evaluates and averages $10$ patches. 
Furthermore, combining it with $10$ common crops achieves the best results for both networks, 
even without using $3$-scale jittering. One interpretation 
for these improvements is that the concentric samples serve a natural prior for the majority of ILSVRC images, 
\ie, the object of interest lies most probably at the center than at the image boundaries. 
This is a common assumption in the literature that also appears in large-scale video segmentation \cite{karpathyTSLSF14}.

Following, we introduce the adaptive sampling mechanism with Algorithm \ref{alg1} and reduce the top-5 error to $15.83\%$ and 
$8.01\%$ for AlexNet and VGG16 respectively. To set this in perspective, Krizhevsky et al. \cite{krizhevskySH12} 
report $16.4\%$ top-5 error when they combine 5 models. We improve this performance with one single model. The relative 
improvement for the deployed instances of AlexNet and VGG16, 
compared to the data-augmentation methods used in \cite{simonyanZ14, szegedyLJSAEVR14}, is $9.9\%$ and $9.4\%$, respectively. 
Row $14$ shows results where the marginalization is weighted based on the entropy (notated as $W$), while the methods in rows $9$--$13$ use uniform weights (\cf Algorithm \ref{alg1}). At the last row we show results from the ILSVRC test server for our top-performing method (row $13$).

%In the Supp. Material we also show results with the GoogLeNet, where our algorithm achieves $21.8\%$ error reduction compared to deploying the model on the whole image and $9.6\%$ compared to $50$ crops at $3$ scales.

\begin{figure}[t]
  \centerline{\includegraphics[width=8.2cm]{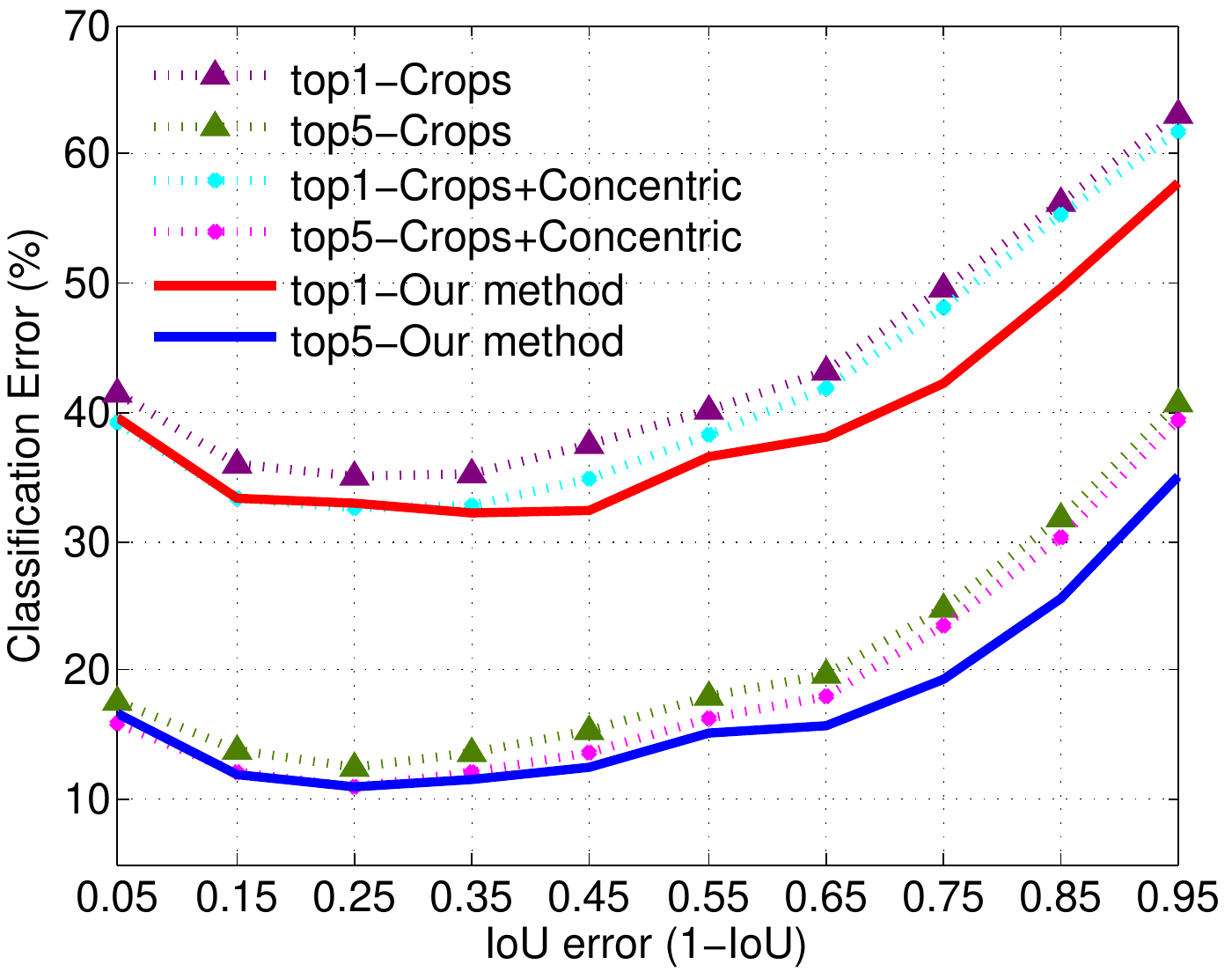}}
\caption{Classification error as a function of the IoU error between the objects and the regular and concentric crops.}
\label{error_vs_iou}
\end{figure}

Regular and concentric crops assume that objects occupy most of the image or appear near the center. This is a known bias in the ImageNet dataset. 
To analyze the effect of adaptive sampling, we calculate the intersection over union error between the objects and the regular and concentric crops, and show 
in Fig.~\ref{error_vs_iou} the performance of various methods as a function of the IoU error. The improvement of using adaptive sampling (via proposals) over only regular and 
concentric crops is increased as IoU error grows, indicating that objects occupy less domain or are far away from the center.

\vspace{-3mm}

\paragraph{Time complexity.}
In Table \ref{table_imagenet} we show the number of evaluated samples ($\#eval$) and the subset that is actually averaged ($\#ave$) 
to extract a single class posterior vector. The sequential time needed for each method is linear to the number of evaluated patches $\#eval$. 
We run the experiments with the MatConvNet library and parallelize the load for VGG16 so that the testing is done 
in batches of $B=20$ patches. We report the time profile\footnote{We use a machine equipped with a NVIDIA Tesla K80 GPU, 24 Intel Xeon E5 cores and 64G RAM memory.} 
for each method in Table \ref{table_imagenet}. A few entries cover two boxes, as their methods are evaluated together. 
Extracting the proposals is not a major bottleneck if using an efficient algorithm \cite{hosangBDS15}, such as Edge Boxes \cite{zitnickD14}.
In rows $13$--$14$ we report results of our faster version, where the Edge Boxes do not leverage edge sharpening and use one decision tree. 
Overall, compared to the $150$-crop strategy, the object proposal scheme introduces marginal computational overhead.\\

\begin{figure*}[t]
\begin{minipage}{0.33\linewidth}
  \centerline{\includegraphics[width=5.5cm]{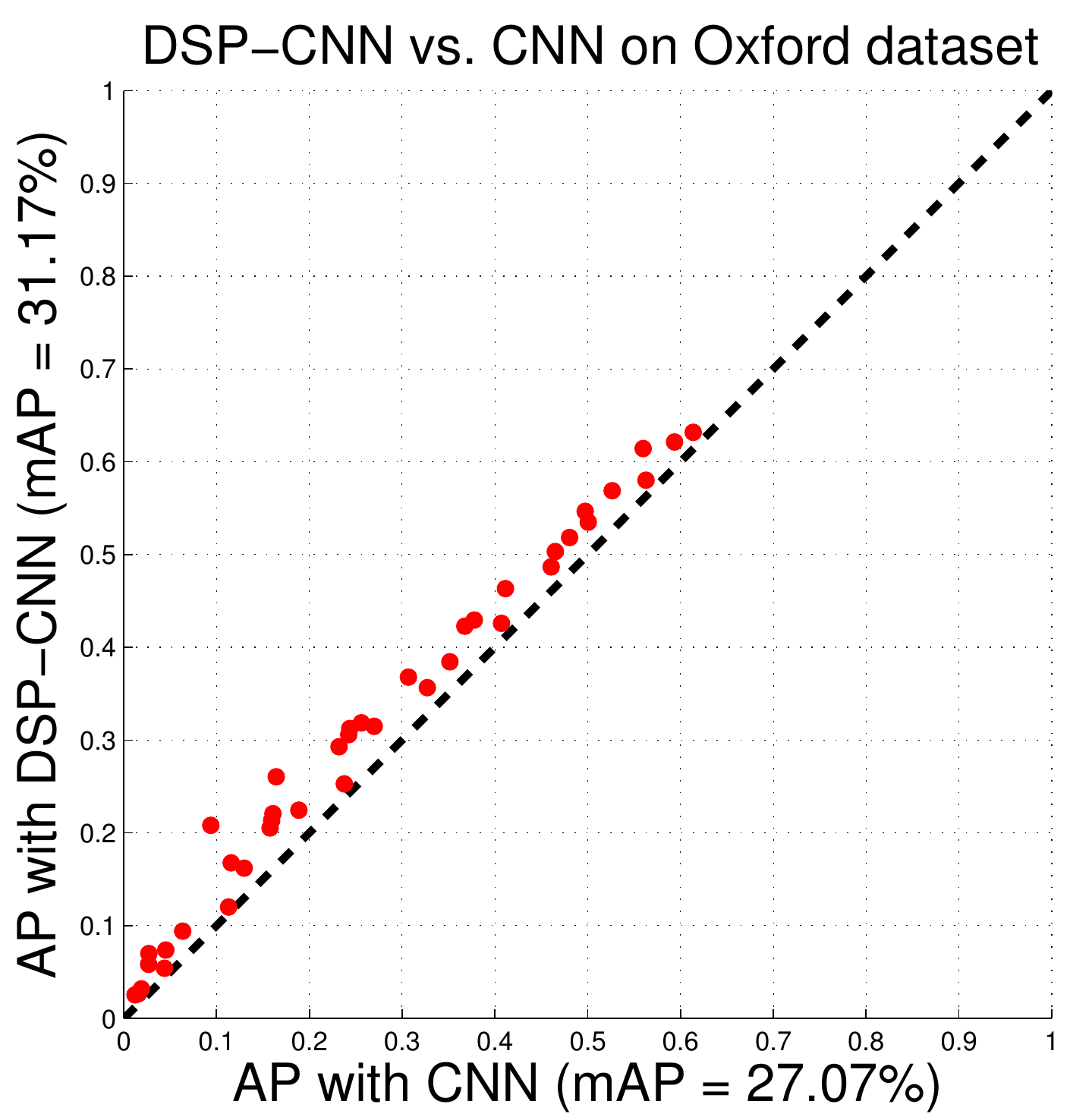}}
\end{minipage}
\begin{minipage}{0.33\linewidth}
  \centerline{\includegraphics[width=5.5cm]{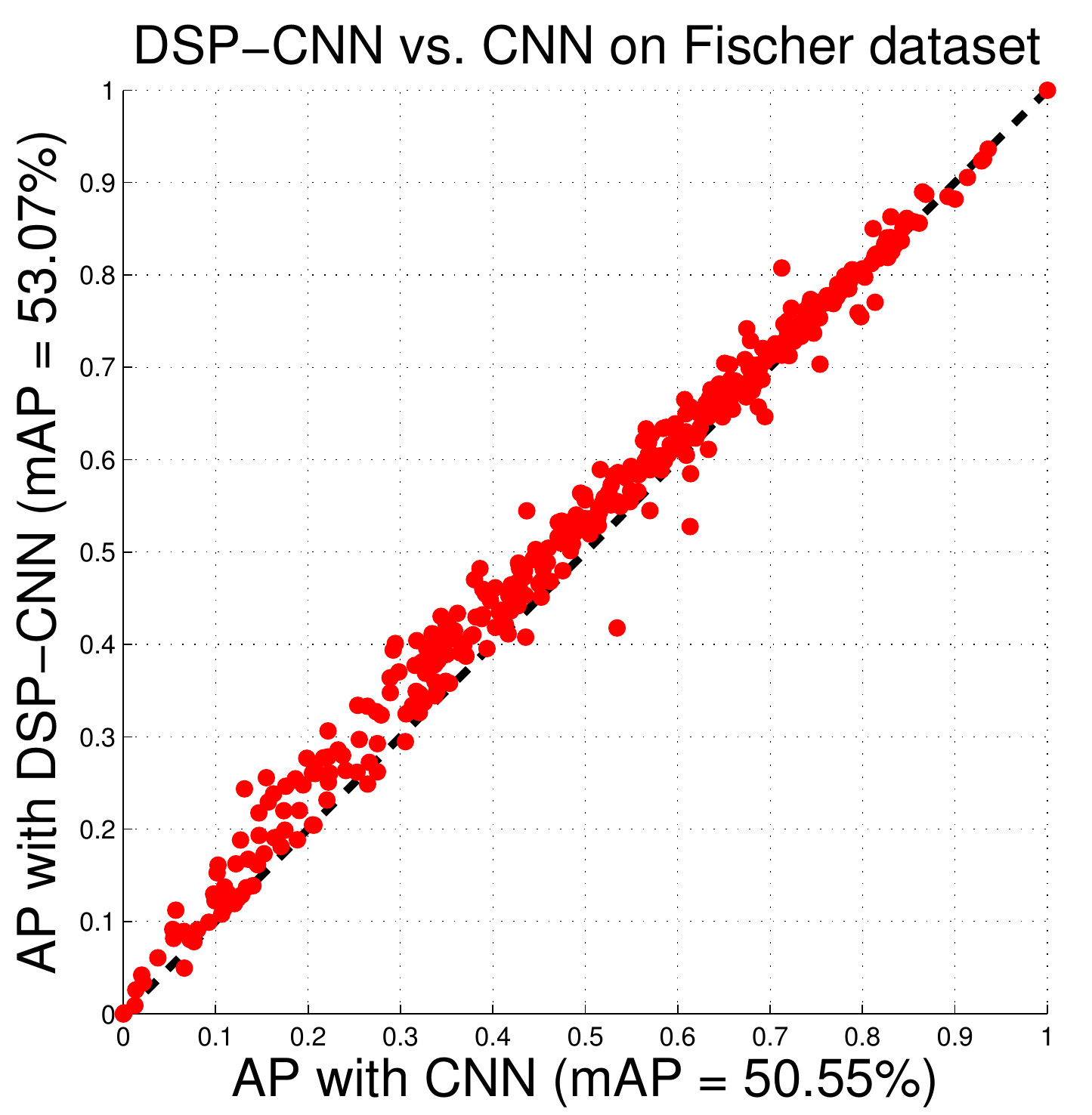}}
\end{minipage}
\begin{minipage}{0.33\linewidth}
  \centerline{\includegraphics[width=5.5cm]{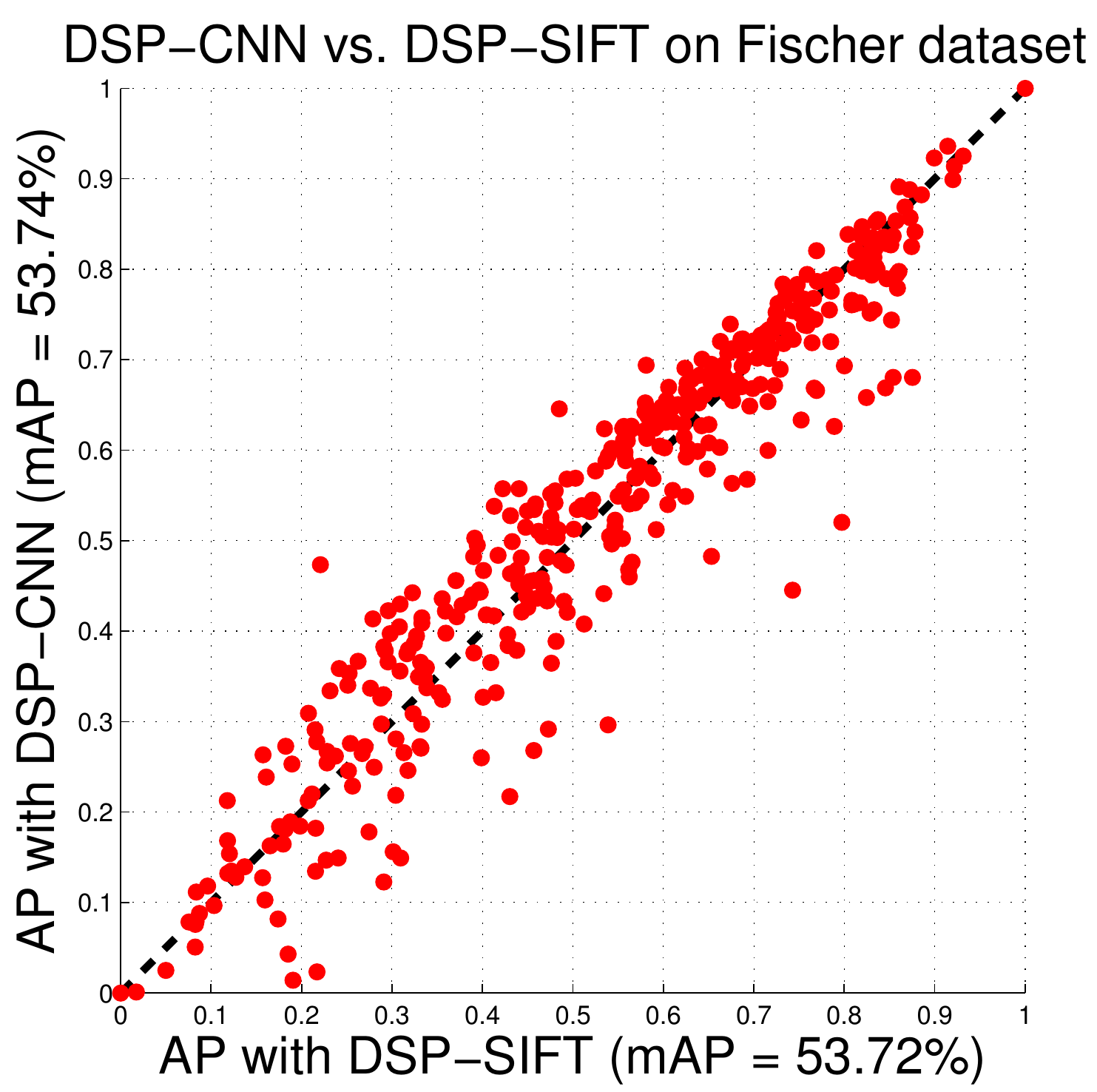}}
\end{minipage}
\vspace{1mm}
\caption{Head to head comparison between CNN and DSP-CNN on the Oxford \cite{mikolajczykTSZMSKG05} (left) and Fischer's \cite{fischerDB14} (center) datasets. 
The layer-4 features of the unsupervised network from \cite{fischerDB14} are used as descriptors. The DSP-CNN outperforms 
its CNN counterpart in terms of matching mAP by $15.1\%$ and $5.0\%$, respectively. Right: DSP-CNN performs comparably 
to the state-of-the-art DSP-SIFT descriptor \cite{dongS15}.}
\label{fig_matching}
\end{figure*}

\vspace{-3mm}

\subsection{Wide-Baseline Correspondence}
\label{subsection-matching}

We test the effect of domain-size pooling in correspondence tasks with a convolutional architecture, as done by \cite{dongS15} for 
SIFT \cite{lowe04}, using the datasets and protocols of \cite{fischerDB14}. This is illustrated in Fig.~\ref{sampling_methods} (upper right), 
but here the domain sizes are centered around the detector. We expect that such averaging will increase the discriminability of
detected regions and in turn the matching ability, similar to the benefits that we see on the last rows of Table \ref{table_gt}.

We use maximally-stable extremal regions (MSER) \cite{matasCUP04} to detect candidate regions, 
affine-normalize them, align them to the dominant orientation, and re-scale them for head-to-head comparisons. 
For a detected scale $\sigma$ at each MSER, the DSP-CNN samples $D$ domain sizes within a neighborhood 
$\lbrack \lambda_1 \sigma, \lambda_2 \sigma \rbrack$ around it, computes the CNN responses on these samples and averages the posteriors. 
The deployed deep network is the unsupervised convolutional network proposed by \cite{fischerDB14}, 
which is trained with surrogate labels from an unlabeled dataset (see the methodology in \cite{dosovitskiySB14}), 
with the objective of being invariant to several transformations that are commonly observed in images captured 
from different viewpoints. As opposed to network-classifiers, here the task is correspondence and the network 
is purely a region descriptor, whose last two layers ($3$ and $4$) are the representations.

\begin{table}[b]
\begin{center}
\begin{tabular}{| c || c | c | }
\hline Method  & Dim & mAP \\ \hline \hline
Raw patch & 4,761 & 34.79\\ \hline
SIFT \cite{lowe04} & 128 & 45.32\\ \hline
DSP-SIFT \cite{dongS15} & 128 & {\bf 53.72}\\ \hline
CNN-L3 \cite{fischerDB14} & 9,216 & 48.99\\ \hline
CNN-L4 \cite{fischerDB14} & 8,192 & 50.55\\ \hline \hline
DSP-CNN-L3 & 9,216 & 52.76\\ \hline
DSP-CNN-L4 & 8,192 & 53.07\\ \hline
DSP-CNN-L3-L4 & 17,408 & {\bf 53.74}\\ \hline \hline
DSP-CNN-L3 (PCA128) & 128 & 51.45\\ \hline
DSP-CNN-L4 (PCA128) & 128 & 52.33\\ \hline
DSP-CNN-L34 (concat. PCA128) & 256 & {\bf 52.69}\\ \hline
\end{tabular}
\end{center}
\caption{\sl \small Matching mean average precision for different approaches on Fischer's dataset \cite{fischerDB14}.}
\label{table_matching}
\end{table}

In Fig.~\ref{fig_matching} (left) we show the comparison between CNN and DSP-CNN on Oxford dataset 
\cite{mikolajczykTSZMSKG05}. CNN's layer 4 is the representation for each MSER and DSP-CNN simply averages 
this layer's responses for all $D$ domain sizes. We use $\lambda_1 = 0.7$, $\lambda_2 = 1.5$ and $D=6$ sizes that are 
uniformly sampled in this neighborhood. There is a $15.1\%$ improvement based on the matching mean average precision.

Fischer's dataset \cite{fischerDB14} includes $400$ pairs of images, some of them with more extreme transformations 
than those in the Oxford dataset. The types of transformations include zooming, blurring, lighting change, rotation, perspective 
and nonlinear transformations. In Fig.~\ref{fig_matching} (center) and Table \ref{table_matching} we show comparisons between CNN 
and DSP-CNN for layer-3 and 
layer-4 representations and demonstrate $7.7\%$ and $5.0\%$ relative improvement. We use $\lambda_1 = 0.5$, 
$\lambda_2 = 1.4$ and $D=10$ domain sizes. These parameters are selected with cross-validation. In Table \ref{table_matching} 
we show comparisons with baselines, such as using the raw data and DSP-SIFT \cite{dongS15}. After fine parameter search 
($\lambda_1 = 0.5$, $\lambda_2 = 1.24$) and concatenating the layers $3$ and $4$, 
we achieve state of the art performance as shown in Fig.~\ref{fig_matching} (right), observing though the high dimensionality of this method 
compared to local descriptors.

%In the Supp. Mat. we report matching performance for varying severity of transformations.

Given the inherent high-dimensionality of CNN layers, we perform dimensionality reduction with principal component 
analysis to investigate how this affects the matching performance. In Table \ref{table_matching} 
we show the performance for compressed layer-3 and layer-4 representations with PCA to $128$ dimensions and
their concatenation. There is a modest performance loss, yet the compressed features outperform the single-scale features by a large margin.

\section{Discussion}
\label{sect-discussion}

Our empirical analysis indicates that CNNs, that are designed to be invariant to nuisance variability due to small planar translations -- by virtue of their convolutional architecture and local spatial pooling -- and learned to manage global translation, distance (scale) and shape (aspect ratio) variability by means of large annotated datasets, in practice are less effective than a naive and in theory counter-productive practice of sampling and averaging the conditionals based on an ad-hoc choice of bounding boxes and their corresponding planar translation, scale and aspect ratio.

This has to be taken with the due caveats: First, we have shown the statement empirically for {\em few} choices of network architectures (AlexNet and VGG), trained on {\em particular} datasets that are unlikely to be representative of the complexity of visual scenes (although they may be representative of the same scenes as portrayed in the test set), and with a specific choice of {\em parameters} made by their respective authors, both for the classifier and for the evaluation protocol. To test the hypothesis in the fairest possible setting, we have kept all these choices constant while comparing a CNN trained, in theory, to ``marginalize" the nuisances thus described, with the same applied to bounding boxes provided by a proposal mechanism. To address the arbitrary choice of proposals, we have employed those used in the current state-of-the-art methods, but we have found the results representative of other choices of proposals.

In addition to answering the question posed in the introduction, along the way we have shown that by framing the marginalization of nuisance variables as the averaging of a {\em sub-sampling} of marginal distributions we can leverage of concepts from classical sampling theory to {\em anti-alias} the overall classifier, which leads to a performance improvement both in categorization, as measured in the ImageNet benchmark, and correspondence, as measured in the Oxford and Fischer's matching benchmarks.

Of course, like any universal approximator, a CNN can in principle capture the geometry of the discriminant surface by ``learning away'' nuisance variability, given sufficient resources in terms of layers, number of filters, and number of training samples. So in the abstract sense a CNN {\em can} indeed marginalize out nuisance variability. The analysis conducted show that, at the level of complexity imposed by current architectures and training set, it does so less effectively than ad-hoc averaging of proposal distributions.

This leaves researchers the choice of investing more effort in the design of proposal mechanisms \cite{girshick15, renHGS15}, subtracting duties from the Category CNN downstream, or invest more effort in scaling up the size and efficiency of learning algorithms for general CNNs so as to render the need for a proposal scheme moot.

%A third way is to alter the architecture of the CNN so it locally marginalize group transformations larger than translation, as currently implied by the convolutional architecture, for instance by extending it to the location-scale group, thus rendering the use of ``domain-size pooling'' moot, to similarity or even affine transformations \cite{cohenW14, gensD14}.

\section*{Acknowledgments}
\label{sect-Acknowledgments}

This research is supported by ARO W911NF-15-1-0564/66731-CS, ONR N00014-13-1-034, and AFOSR FA9550-15-1-0229. We gratefully acknowledge NVIDIA Corporation for donating a K40 GPU that was used in support of some of the experiments.

{\small
\bibliographystyle{ieee}
\bibliography{egbib}

}

\end{document}